# SITUATED MODELING OF EPISTEMIC PUZZLES

Murat Ersan and Varol Akman

# BILKENT UNIVERSITY

Department of Computer Engineering
and
Information Science

# SITUATED MODELING OF EPISTEMIC PUZZLES[1]

Murat Ersan and Varol Akman

September, 1994

**Abstract**

Situation theory is a mathematical theory of meaning introduced by Jon Barwise and John Perry. It has evoked great theoretical and practical interest and motivated the framework of a few 'computational' systems. PROSIT is the pioneering work in this direction. Unfortunately, there is a lack of real-life applications on these systems and this study is a preliminary attempt to remedy this deficiency. Here, we examine how much PROSIT reflects situation-theoretic concepts and solve a group of epistemic puzzles using the constructs provided by this programming language.

# Chapter 1

# Introduction

Situation theory is a principled programme to develop a mathematical theory of meaning which aims to clarify and resolve some tough problems in the study of language, information, logic, and philosophy [7]. It was introduced by Jon Barwise and John Perry and stimulated great interest [8]. The theory matured within the last ten years or so [5, 14, 15, 36, 52, 53] and various versions of it have been applied to a number of linguistic issues [17], resulting in what is commonly known as *situation semantics*. This was followed by assorted studies on the computational aspects of the theory, which gave birth to a group of computational systems based on situation theory [32, 33, 35, 44, 48, 46, 50, 49, 9, 10].

PROSIT (PROgramming in SItuation Theory), developed by Nakashima et al. [32, 33, 35], is *the* pioneering work in this direction. Therefore, it is worth examining how much PROSIT reflects situation-theoretic concepts and how much it deviates from them. PROSIT seems to be especially suitable for writing programs simulating humanlike (commonsense) reasoning [28, 29]. Unfortunately, there have been very few attempts to employ PROSIT in this style. Such a study is, however, of great importance, and would help us see where and why we should utilize systems based on situation theory, and how we should go about formulating a situation-theoretic programming paradigm [48, 46]. In fact, as far as we know, the only remarkable application in which PROSIT has been effectively exploited is the "Three Wisemen Problem" [32]. This is a problem involving common knowledge (mutual information) in a multi-agent setting. Pinning our faith upon situation theory, we tried to make use of PROSIT in the solution of what we came to call 'epistemic puzzles' [19, 26, 30, 38, 39, 40, 41, 42]. Throughout this paper the nature of epistemic puzzles and their solution via a situation-theoretic world-view will be analyzed.

A short introduction to situation theory and situation semantics, and two other computational systems (ASTL and BABY-SIT) will be offered in the next two chapters, respectively. This is followed by a detailed review of PROSIT, where a section is devoted to the comparison of PROSIT and bona fide situation theory. The fifth chapter explains what epistemic puzzles look like, how they have been solved using classical approaches, and why the situated model fits best to model and solve these puzzles. The discussion is supported with a great variety of puzzles, some of which are introduced by Raymond Smullyan in his book *Forever Undecided: A Puzzle Guide to Gödel* [42]. (Also cf. [38, 39, 40] for similar puzzles.) The discussion ends with a conclusion and proposed future work.

# Chapter 2

# Situation Theory and Situation Semantics

Situation theory is a mathematical theory of meaning. It was introduced by Barwise and Perry in their book *Situations and Attitudes* [8] and evoked great theoretical and practical interest.

Barwise and Perry were aware of the limitations of classical logic and contended that the standard view of logic is inappropriate for many of the uses to which it has been put by semanticists. There have been different approaches to building theories of natural language. Some of these theories emphasized the power of language to classify minds, i.e., the mental significance of language, while others focused on the connections between language and the described world, i.e., the external significance of language. However, Barwise and Perry claim that for an expression to have meaning, it should convey information. This is possible, only if the expressions have a link with the kinds of events they describe and also a link with the states of mind. They develop a theory of situations and of meaning as a relation between situations. The theory provides a system of abstract objects that help describe the meaning of both expressions and mental states in terms of the information they carry about the external world.

Keith Devlin [17], who spared considerable effort on the formalization of the theory, also regards situation theory as a theory of information. Rather than try to define information, he investigated the nature of information flow [18] and the mechanisms that gave rise to such flow.

The evolution of situation theory can be regarded as a move away from conventional logics which only have relatively simple objects in the semantic domain to more complex semantic objects. Within this movement, although at the beginning there was little distinction, today there is a split between situation theory—the formal aspects of the theory, such as mathematical, logical, philosophical, proof theoretic, etc.—and situation semantics—the application of situation theory to the semantics of natural language.

The information-based approach to the semantics of natural languages has resulted in what is known as situation semantics. The primary idea situation semantics is based on is that language is used to convey information about the world. Two sentences with the same interpretation—describing the same situation—can carry different information. Context-dependence, which was underestimated in classical approaches to semantics, is the essential hypothesis of situation semantics. Indexicals, demonstratives, tenses, and other linguistic devices rely heavily on context for their interpretation [1]. Therefore, a sentence can be used over and over again in different situations to say different things. Its interpretation is subordinate to the situation in which the sentence is uttered.

The framework of situation theory mainly consists of the things an (intelligent) agent is able to discriminate using his cognitive abilities. The basic ingredients of this framework are

- *Individuals*, which are considered as entities that are individuated as 'objects'
- *Properties* that hold or fail to hold for some of these individuals
- *Relations* that hold or fail to hold among some of those individuals
- *Spatial* and *temporal locations* that are points in regions of space and time.

The two major notions of situation theory are *infons* and *situations*. Infons are the basic informational units. They should be considered as discrete items of information. Infons are denoted as $\ll P, a_1, \ldots a_n, i \gg$ where $P$ is an $n$-place relation, $a_1, \ldots a_n$ are objects appropriate for the respective *argument places* of $P$, and $i$ is the *polarity* (0 or 1). It is possible to use spatial and temporal locations in the argument places of relations. The following infon states the fact that Bob is married to Carol at time t:

$\ll$married-to,Bob,Carol,t,1$\gg$

Situations are 'first-class' citizens of the theory. There is no clear-cut definition of what a situation exactly is. Rather, a situation is considered to be a structured part of the Reality that a cognitive agent somehow manages to pick out (individuate). Situations *support* facts:

*s* supports $\alpha$ ($s \models \alpha$) means that $\alpha$ is an infon that is true of situation *s*.

A simple example would be

$S_1 \models \ll$running,Bob,1$\gg$

which states that Bob is running in situation $S_1$.

One should note that the truth or falsity of a fact does not depend on the *supports* relation but is handled by the notion of polarity. Therefore

$S_1 \not\models \ll$running,Bob,1$\gg$

does not imply

$S_1 \models \ll$running,Bob,0$\gg$

One of the primary motivations for situation theory was sentences of the form

Bob is angry.
Bob is angry and Bob is shouting or Bob is not shouting.

In conventional classical logic there is no way to distinguish between these two sentences. They are considered to be logically equivalent, because any truth assignment that makes the former true will also make the latter true. However, intuitively there seems to be a difference. In situation theory situations are *partial*, i.e., they do not define the truth or falsity of all relations on all objects in the domain. Permitting partiality, the theory can distinguish between those sentences that look logically equivalent. In situation semantics these two sentences will have different interpretations. The first one will be represented by a situation, $S$, in which Bob is angry. $S$ does not state anything about (is not aware of) Bob's shouting or not. Another situation, $S'$, represents the second sentence. $S'$ is the 'union' of two situations: the situation in which Bob is angry and shouting, and the situation in which Bob is angry and not shouting.

It is desirable to have some computational tools to handle situations. Abstract situations are the mathematical constructs with which we can model analogs of real situations. They are more amenable to mathematical manipulation. An abstract situation is defined to be a set of infons. Given a real situation $s$ the set $\{\alpha \mid s \models \alpha\}$ is the corresponding abstract situation.

An important feature of situation theory is the existence of types. Types are higher-order uniformities which cut across individuals, relations, situations, and spatial and temporal locations. Just as individuals, temporal locations, spatial locations, relations, and situations, types are also discriminated by cognitive agents. In this framework, relations may have their argument places filled either with individuals, situations, locations, and other relations or with types of individuals, situations, locations, and relations. For example, if an agent sees smoke he can conclude that there is fire. For he is aware of the constraint which links situations where there is smoke to those where there is fire. Thus, the constraint links types of situations, viz. smoky-type of situations to ones with fire.

The development of types brings the requirement of devices for making reference to arbitrary objects of a given type. Therefore for each type $T$, an infinite collection of *parameters* $T_1, T_2, \ldots$ is introduced. For example $IND_3$ is an $IND$-parameter (parameters of type $IND$). Given a $SIT$-parameter, $SIT_i$, and a set of infons, $I$, the following denotes a situation-type, the type of situation in which conditions in $I$ are satisfied:

$$[SIT_i \mid SIT_i \models I]$$

For example,

$$[SIT_1 \mid SIT_1 \models \ll\text{running}, \text{Bob}, LOC_1, TIM_1, 1\gg]$$

denotes the type of situation in which Bob is running at some location and at some time. These parameters already carry some computational power, but we need more than that. Rather than parameters ranging over all individuals, we need parameters that range over a more limited class. Such parameters are called *restricted parameters*. Given a basic parameter, $v$, and a condition, $C$, on $v$, a restricted parameter $v \upharpoonright C$ is defined. This is of the same basic type as $v$ and satisfies the requirements imposed by $C$. For example,

$$\dot{b} = IND_2 \upharpoonright \ll\text{football}, IND_2, 1\gg$$
$$\dot{a} = IND_3 \upharpoonright \{\ll\text{man}, IND_3, 1\gg, \ll\text{kicking}, IND_3, \dot{b}, 1\gg\}$$

Once defined, $\dot{b}$ ranges over all footballs and $\dot{a}$ over all men kicking footballs.

In addition, it is possible to obtain new types using a parameter, $s$, and a set, $I$, of infons (in the form $[s \mid s \models I]$). For example,

$$[SIT_1 \mid SIT_1 \models \ll\text{kicking}, \dot{a}, \dot{b}, 1\gg]$$

represents a situation-type where a man is kicking a football and

$$[\dot{a} \mid SIT_1 \models \ll\text{kicking}, \dot{a}, \dot{b}, 1\gg]$$

denotes the type of men kicking a football.

In situation theory, the flow of information is realized by a certain group of infons called *constraints*. A situation $s$ will carry information relative to the constraint $C = [S \Rightarrow S']$, if $s : S[f]$, where $f$ *anchors* the parameters in $S$ and $S'$. Hence, the information carried by $s$ relative to $C$ is that there is a situation $s'$, possibly extending $s$, of type $S'[f]$.

This introduction on situation theory and situation semantics will be finished by a previous example about constraints.

$$S_0 = [\dot{s_0} \mid \dot{s_0} \models \ll\text{smoke-present}, \dot{l}, \dot{t}, 1\gg]$$
$$S_1 = [\dot{s_1} \mid \dot{s_1} \models \ll\text{fire-present}, \dot{l}, \dot{t}, 1\gg]$$
$$C = [S_0 \Rightarrow S_1],$$

In this example, $C$ is a constraint that links situations where there is fire, $S_1$, to situations where there is smoke, $S_0$. An agent who is aware of (attuned to) this constraint will infer that there is a fire whenever he perceives smoke.

# Chapter 3

# Computational Situation Theory

Currently, there are three systems based on situation theory. PROSIT, developed by Nakashima et al. [32, 33, 35], is *the* pioneering work in this direction. This was followed by the development of ASTL by Black [9, 10]. Another computational medium based on situations called BABY-SIT is currently being built at Bilkent University by Akman and Tın [44, 48, 46]. PROSIT is primarily aimed at general problems of knowledge representation, while ASTL is developed for experiments in natural language processing. On the other hand, BABY-SIT will hopefully handle problems of both sorts. In the following sections brief explanations of ASTL and BABY-SIT will be given. Because the epistemic puzzles in this paper are implemented on PROSIT, a separate chapter is devoted to explain that language.

## 3.1 ASTL

ASTL (A Situation Theoretic Language) is a situation theoretic language developed by Alan Black in 1991. It was primarily designed to make experiments on semantic theories of natural language. Black chose situation theory as a basis for his system because it provides a general and potentially powerful formalism on which theories of natural language processing could be implemented. The system consists of an interpreter (implemented in Common Lisp) that passes over the ASTL definitions to make inference and answer queries about a set of constraints and basic situations.

### 3.1.1 Syntax and Semantics of ASTL

Similar to many logic programming languages, ASTL has a syntax that consists of terms and sentences. Terms can be simple or complex. Simple terms are used to denote *individuals* (e.g., `a, b`), *relations* (e.g., `happy, sees`), *parameters* (e.g., `X, Y`), and *variables* (e.g., `*P, *R`). Complex terms are built up from simple terms. The complex terms and their notations are as follows.

*I-terms* are used to represent the basic informational units, i.e., infons. The syntax is $\langle\langle rel, arg_1, \ldots, arg_n, polarity \rangle\rangle$ where $rel$ is a relation of arity $n$, $arg_1$ to $arg_n$ are the terms that stand for the arguments of the relation, and *polarity* denotes the polarity of the infon. For example, an infon stating the fact that Bob is singing may be represented by the following i-term:

```
            <<singing,bob,1>>
```

*Types* are complex terms denoting the types of situations. They are represented as [*parameter* ! *condition*$_1$ ... *condition*$_n$] where each condition has the form *parameter* != *i-term*. For example, the following represents a situation type where Bob is happy and Carol is sad.

```
[S ! S != <<happy,bob,1>> S != <<sad,carol,1>>]
```

*Situations* are represented by atomic names which are optionally followed by a type (separated by a double colon). An example for a situation would be

```
SIT1 ::   [S ! S != <<happy,bob,1>> S != <<smiling,bob,1>>]
```

Here `SIT1` is of type `S`, i.e., it is a situation where Bob is happy and smiling.

ASTL sentences are defined using the simple and complex terms. It should be noted however, that sentences are distinct from terms and cannot be used as arguments to relations. The following are some of the basic ASTL sentence types.

*Propositions* are situation names followed by types, separated by a single colon. For example,

```
SIT2 :   [S ! S != <<happy,bob,1>> S != <<happy,carol,1>>]
```

indicates that both Bob and Carol are happy in situation `SIT2`.

*Constraints* are the primary tools that are used in inferencing. They are defined between propositions, and are of the form $sit_0 : type_0$ <= $sit_1 : type_1, \ldots, sit_n : type_n$ where each $sit_i$ is a situation name or a variable and $type_i$ is a type. The following example demonstrates a constraint which states that whenever Bob is smiling, he is happy:

```
*S : [S ! S != <<happy,bob,1>>] <= *S : [S ! S != <<smiling,bob,1>>]
```

*Grammar rules* are a special kind of constraints and have a similar semantics. They are of the form $sit_0 : type_0$ -> $sit_1 : type_1, \ldots, sit_n : type_n$. An example is

```
*S  : [S ! S != <<category,*S,Sentence,1>>] ->
*NP : [S ! S != <<category,*NP,NounPhrase,1>>],
*VP : [S ! S != <<category,*VP,VerbPhrase,1>>].
```

This grammar rule should be interpreted as follows: if there is a situation `*NP` of the given type and a situation `*VP` of the given type then there is also a situation `*S` of the given type.

### 3.1.2 Inference in ASTL

The primary sentences used to make inference are the constraints. ASTL has five inference rules using which it answers non-trivial questions about the described models. The following paragraphs explain these inference rules and offer examples.

*Type reduction* aims to break down a type with more than one condition into pieces. For example the situation

```
SIT1 :   [S ! S != <<happy,bob,1>> S != <<singing,bob,1>>]
```

is also a situation of the following types:

```
SIT1 :   [S ! S != <<happy,bob,1>>]
SIT1 :   [S ! S != <<singing,bob,1>>]
```

*Type combination* is the reverse of type reduction, i.e., it combines single condition propositions of the same situation. For example, if the following propositions are true

```
SIT1 :   [S ! S != <<happy,bob,1>>]
SIT1 :   [S ! S != <<dancing,bob,1>> S != <<singing,bob,1>>]
```

then the following proposition is also true

```
SIT1 :   [S ! S != <<happy,bob,1>> S != <<dancing,bob,1>> S != <<singing,bob,1>>]
```

*Modus ponens* derives the conclusion of a constraint if the premise(s) is (are) satisfied. So, if we have a constraint and a proposition such as

```
SIT1 :   [S ! S != <<happy,bob,1>>] <= SIT2 :   [S ! S != <<singing,bob,1>>]

SIT2 :   [S ! S != <<singing,bob,1>>]
```

then ASTL deduces the following proposition

```
SIT1 :   [S ! S != <<happy,bob,1>>]
```

*Argument promotion* allows the proper treatment of typed situations as arguments in facts. For example, using the following constraint and proposition

```
SIT0 :   [S ! S != <<happy,bob,1>>] <=
SIT1 :   [S ! S != <<sees,bob, SIT2 ::   [S ! S != <<happy,carol,1>>],1>>]

SIT1 :   [S ! S != <<sees,bob,SIT2,1>>]
```

it is possible to deduce the following constraint

```
SIT0 :   [S ! S != <<happy,bob,1>>] <= SIT2 :   [S ! S != <<happy,carol,1>>]
```

ASTL also has a mechanism to handle cyclic constraint definitions. It is possible to deduce

```
SIT1 :   [S ! S != <<happy,bob,1>>]
```

using the following constraint and proposition:

```
*S : [S ! S != <<happy,bob,1>>] <=
*S : [S ! S != <<sees,bob,*S,1>> S != <<happy,carol,1>>]

SIT1 :   [S ! S != <<sees,bob,SIT1,1>> S != <<happy,carol,1>>]
```

### 3.1.3 Applications

Sentences are parsed in ASTL according to the situation theoretic grammar (STG) described by Cooper [13], and situation-theoretic representations of them are built. In this grammar, situations are used to represent actual utterances. These are called *utterance situations*. Using utterance situations it is possible to write situation-theoretic constraints. Because ASTL is aimed at natural language processing, it has special constructs to handle utterance situations. Rather than using constraints and fully specifying all connections in an utterance, i.e., the order of words, grammar rules are used. Grammar rules are a special form of constraints which are more compact and are more efficient to use.

For example, the following constraint states that, if there is a situation that supports the fact that it is a noun phrase, and a situation that is a verb phrase, and that the noun phrase is connected to a point which the verb phrase starts from (*Mid), then there is a situation which is a sentence:

```
*S : [S ! S != <<category,S,Sentence,1>>
        S != <<connected,S,*Start,*End,1>>
        S != <<uttsit,S,1>>
        S != <<daughters,S,*NP,*VP,1>>]
<=
   *NP : [S ! S != <<category,S,NounPhrase,1>>]
           S != <<uttsit,S,1>>
           S != <<connected,S,*Start,*Mid,1>>
   *VP : [S ! S != <<category,S,VerbPhrase,1>>]
           S != <<uttsit,S,1>>
           S != <<connected,S,*Mid,*End,1>>
```

On the other hand, using grammar rules it is easier to represent the same constraint as,

```
*S : [S ! S != <<category,S,Sentence,1>>
        S != <<daughters,S,*NP,*VP,1>>]
->
   *NP : [S ! S != <<category,S,NounPhrase,1>>]
   *VP : [S ! S != <<category,S,VerbPhrase,1>>]
```

ASTL has been used as a meta-language for several important semantic theories. One of these semantic theories is the Discourse Representation Theory (DRT) by Kamp [25], that introduces the notion of *discourse representation structure* which can be considered as representing states in the discourse. Another one is the Dynamic Predicate Logic (DPL) [20] which views semantics of an expression as a relation between an input state and an output state. These semantic theories have been implemented and have been compared using the constructs of ASTL. Although ASTL reflects a small part of situation theory, i.e., infons, situations, and constraints, it has been successful enough to offer useful ideas.

## 3.2 BABY-SIT

BABY-SIT is a computational medium based on situations. It is currently being developed at Bilkent University in KEE$^{TM}$ (Knowledge Engineering Environment) on a SPARCstation$^{TM}$. BABY-SIT is mainly aimed at developing and testing programs in domains ranging from linguistics to artificial intelligence within a framework built upon situation theoretic constructs.

### 3.2.1 Syntax and Semantics

The computational model of BABY-SIT consists of nine primitive domains: individuals, times, places, relations, polarities, parameters, infons, situations, and types. Each of these primitive domains has its own internal structure. *Individuals* are unique atomic entities. They correspond to the objects in the world. *Times* are used to represent temporal locations and are a distinguished type of individuals. Similar to times, *places* are individuals that represent spatial locations. *Relations* correspond to relations that hold between objects in the world. They have certain argument roles which must be occupied by appropriate objects. *Polarities* represent the truth values of relations. *Infons* are discrete items of information and are of the form <<$rel, arg_1, \ldots, arg_n, pol$>>, where $rel$ is a relation, $arg_i, 1 \leq i \leq n$ is an object of the appropriate type for the $i$th argument role, and $pol$ is the polarity. *Parameters* are 'place holders' of the objects in the model. Using parameters one can refer to arbitrary objects of a given type. (Abstract) *situations* are sets of parametric infons. *Types* are higher-order uniformities that individuate (discriminate) objects in the world. BABY-SIT offers nine primitive types: ~IND (individuals), ~TIM (times), ~LOC (places), ~REL (relations), ~POL (polarities), ~INF (infons), ~PAR (parameters), ~SIT (situations), and ~TYP (types).

BABY-SIT offers two modes of interaction with the system. The assertion mode provides an interactive environment where one can define objects and their types, and assert infons into situations. The query mode enables one to issue queries about the existing situations.

In the following example, it is asserted that `bob` and `mary` are individuals, `loves` is a relation, and `sit1` is a situation. `I>` is the prompt of the assertion mode:

```
I> bob:    ~IND
I> mary:   ~IND
I> loves:  ~REL
I> sit1:   ~SIT
```

To state the fact that Bob loves Mary in a situation the following assertion is made:

```
I> sit1 |= <<loves,bob,mary,1>>
```

Queries are handled by either direct querying through situations, i.e., using the existing infons in situations, or by the application of backward-chaining constraints (explained in the next section). BABY-SIT offers different types of queries that can be controlled by the user:

- Searching for solutions by using a given group of constraints.

- Replacing each parameter in the query expression by the corresponding individual if there is a possible anchor.

- Returning a specified number of solutions.

- Displaying solutions with the parameters replaced with the corresponding individuals.

- Displaying a trace of the anchoring of parameters in each solution.

A simple query asking for the person Bob loves would be

    Q> sit1 |= <<loves,bob,?X,1>>

to which the system will respond

    sit1 |= <<loves,bob,mary,1>>

### 3.2.2 Inference

In BABY-SIT, inference is made via constraints. Constraints can be forward-chaining, backward-chaining, and both forward- and backward-chaining. Each constraint has an identifier and is a member of a group of constraints. The following constraint has the identifier BEING and is a member of the constraint group BEING-PERSPECTIVE. It states that every man is a human.

```
BEING-PERSPECTIVE:
BEING:
?S |= <<human,?X,1>>
<=
   ?S |= <<man,?X,1>>
```

Constraints can be either global or situated; i.e., they can be applied to all situations or to a specific one. It is also possible to add some *background constraints* which must be satisfied for the constraint to apply. For example, the following constraint states that balls fall if they are not supported, but only if there is gravity:

```
NATURAL-LAW-PERSPECTIVE:
FALLING-BALL:
?S1 |= <<ball,?X,1>>,
?S1 |= <<supported,?X,0>>
=>
  ?S2 |= <<falls,?X,1>>
UNDER-CONDITIONS:
w:  {<<exists,gravity,1>>}
```

(w denotes the background situation.) This is the situation from which all the other real-life situations normally inherit.

Forward chaining constraints are activated whenever their antecedents are satisfied. All the consequences are then asserted. New assertions may in turn activate other forward chaining constraints. A backward chaining constraint is activated when a query is made about its consequence. The system then tries to satisfy all the antecedents of the constraint.

### 3.2.3 Applications

BABY-SIT has been used to resolve pronominal anaphoric expressions in Turkish [50]. This process can be defined as finding the antecedent and referent of an anaphoric expression. Consider the following examples:

(1) Bilge bana [∅ hastalandığın]-ı söyledi.
(2) Erol maça gelmeyecek. Bilge bana [∅ hastalandığın]-ı söyledi.

In the first example, the zero anaphor expression, ∅, being the subject of the embedded sentence will take the subject of the main sentence, Bilge, as its antecedent. However, in the second example, the first sentence supplying the context, the antecedent of the zero anaphor is the subject of the previous sentence, i.e., Erol.

As a starting point, the existing syntactic approaches to resolve pronominal anaphora for isolated sentences has been implemented in BABY-SIT. This was followed by the generation of simple syntactic rules to resolve the issue across sentence boundaries. An example for such a rule would be: "if the subject of the main sentence is represented by a zero pronoun, then it co-refers with the subject of the immediately preceding sentence."

The procedure BABY-SIT follows in resolving anaphora can be summarized as follows. Each linguistic expression is considered as an utterance situation. Therefore, for each linguistic expression in a sentence type of utterance situation is defined. The situation that represents the whole sentence is designated by the composition of the situations of its sub-utterances. There is also a group of constraints that place restriction on the existing environment. An example for such a constraint would be: "if there is an utterance situation of the word 'AYNUR', then it must represent a female human being." Additionally, there is a background situation that contains information about the utterer and the addressee. After all the utterance situations are asserted and the constraints are satisfied, rules that encode syntactic control of zero anaphora are exploited to resolve the anaphora.

BABY-SIT has also been used to implement the causal theories of Shoham [37], and their extensions proposed by Tın and Akman [45]. These theories were tested on a group of problems, one of which is the famous *Yale Shooting Problem* (YSP). This is a puzzle proposed by Hanks and McDermott [21] as a paradigm to show how the temporal projection problem arises:

> At some point in time, a person, Fred, is alive. A loaded gun, after waiting for a while, is fired at Fred. What are the results of this action?

One expects that Fred would die and that the gun would be unloaded after the firing. But Hanks and McDermott [22] demonstrate in the framework of (among other formalisms) circumscription, that unintended minimal models are obtained; the gun gets unloaded during the waiting stage and firing the gun does not kill Fred.

Causal theories try to reason about the effects of such actions. Proceeding in time, the causal inference mechanism tries to obtain knowledge about future using what is known (and what is not known) about the past. The axioms of causal theories are translated into infons and constraints in BABY-SIT. For example, the following constraint is, in fact, the translation of the axiom stating that if a gun is loaded at some point in time it will continue to be loaded unless someone fires it or manually empties it:

```
GUNFIRE:
R3:
?S1 |= <<loaded,?G,1>>
=>
   ?S3 |= <<loaded,?G,1>>,
   ?S1 |= <<successor,?S3,?S1,1>>
UNDER-CONDITIONS:
?S1:  {<<fires,?M,?G,0>>,
       <<emptied-manually,?G,0>>,
       <<successor,?S2,?S1,0>>}
```

Both Shoham's causal theories, and Tın and Akman's [47] extensions that permit simultaneity have been successfully modeled and compared in BABY-SIT.

# Chapter 4

# PROSIT

PROSIT [32, 33, 35] is a declarative language in which both programs and data are just sets of *infons*. This feature makes PROSIT akin to Prolog, but PROSIT is based on situation theory rather than Horn clauses. The motivation behind the design of this language rests on the following desirable features, each of which is supported by the theory:

- The use of partially specified objects and partial information
- Situations as first-class citizens
- Situatedness of information and constraints
- Informational constraints
- Self-referential expressions

These features provide the necessary power to analyze semantic phenomena in natural language. PROSIT also offers tools for knowledge representation, interactive querying, and deduction, which are important components of a programming environment.

## 4.1 Syntax and Semantics

Expressions in PROSIT are either atoms or lists. Atoms that are numbers (2, 3.5, 527, etc.) or strings (``hello'', ``enter'', etc.) are considered to be *constants*, whereas atoms that are symbols (FOO, *A, B44, etc.) are regarded either as *parameters* or *variables*. Lists are similar to Lisp lists, i.e., they are a series of atoms or lists separated by spaces and enclosed by parentheses, as in (A (B C (D)) E).

Parameters are Lisp symbols starting with a character other than "*". They are used to represent things that cannot be captured by PROSIT constants, such as objects, situations, and relations. Usually, different parameters correspond to different entities. Parameters can be used in any infon (including queries and constraints); their scope is global.

Symbols starting with "*" are variables. Variables are place-holders that stand for any PROSIT expression. They only appear in queries and constraints; their scope is local to the constraint or query they participate in. If a variable is bound to a certain value, then in the later parts of the same constraint or query, those variables are replaced with their value unless the system backtracks.

In PROSIT, an infon is represented as a list whose first element is the symbol for a relation and whose remaining elements are the objects for which the relation holds:

(*relation object$_1$ ... object$_n$*)

For example, the infon

    (listening_to John Mary)

expresses that the relation `listening_to` holds between the objects represented by the parameters `John` and `Mary`, i.e., John is listening to Mary.

One can assert infons, and query a knowledge base incorporating, among other things, infons. Unlike Prolog, all infons are local to situations. For example, to assert the infon mentioned above into a situation, `sit1`, the following expression is used:

    (!= sit1 (listening_to John Mary))

In PROSIT, there exists a tree hierarchy among all situations, with the situation `top` at the root of the tree. `top` is the global situation and the 'owner' of all the other situations generated. One can traverse the 'situation tree' using the predicates `in` and `out`. Although it is possible to issue queries from any situation about any other situation, the result will depend on where the query is made. If a situation `sit2` is defined in the current situation, say `sit1`, then `sit1` is said to be the owner of `sit2`, or equivalently:

- `sit2` is a part of `sit1`, or
- `sit1` describes `sit2`

The owner relation states that if `(!= sit2 infon)` holds in `sit1`, then `infon` holds in `sit2`, and conversely, if `infon` holds in `sit2` then `(!= sit2 infon)` holds in `sit1`.

`in` causes the interpreter to go to a specified situation which will be a part of the 'current situation' (the situation in which the predicate is called). `out` causes the interpreter to go to the owner of the current situation.

Similar to the owner relation there is the 'subchunk' relation, denoted by `([_ sit1 sit2)`, where `sit1` is a subchunk of `sit2`, and conversely, `sit2` is a 'superchunk' of `sit1`. When `sit1` is asserted to be the subchunk of `sit2` it means that `sit1` is totally described by `sit2`. A superchunk is like an owner (except that `out` will always cause the interpreter to go to the owner, not to a superchunk).

PROSIT has two more relations that can be defined between situations. These are the 'subtype' and the 'subsituation' relations. When the subtype relation, denoted by `(@< sit1 sit2)`, is asserted, it causes the current situation to describe that `sit2` supports each infon valid in `sit1` and that `sit2` respects every constraint that is respected by `sit1`, i.e., `sit1` becomes a subtype of `sit2`. The subsituation relation, denoted by `(<-- sit1 sit2)`, is the same as `(@< sit1 sit2)` except that only infons, but no constraints, are inherited. Both relations are transitive.

A distinguishing feature of PROSIT is that the language allows circularity [6]. The fact that PROSIT permits situations as arguments to infons makes it possible to write self-referential statements. Consider a card game (`sit`) between two players. John has the ace of spades and Mary has the queen of spades. When both players display their cards the following infons will be factual:

```
(!= sit (has John ace_of_spades))
(!= sit (has Mary queen_of_spades))
(!= sit (sees John sit))
(!= sit (sees Mary sit))
```

In this example the third and the fourth infons are circular, viz. `sit` supports facts in which it appears as an argument.

## 4.2 Inference

The notion of informational constraints is a distinguishing feature that shaped the design of PROSIT. Constraints can be considered as special types of information that 'generate' new facts. They are just a special case of infons, and therefore, are also situated. A constraint can be specified using either of the three relations =>, <=, and <=>. Constraints specified with => are forward-chaining. They are of the form (=> *fact head$_1$ head$_2$ ...head$_n$*). If *fact* is asserted to the situation then all of the head facts are also asserted to that situation. Constraints specified with <= are backward-chaining. They are of the form (<= *head fact$_1$ fact$_2$ ...fact$_n$*). If each of the facts from 1 to *n* are supported by the situation, then *head* is also supported (though not asserted) by the same situation. Finally, constraints specified with <=> should be considered as both backward- and forward-chaining.

If there is a constraint stating that "everything that smiles is happy" in situation `sit1`,

```
(resp sit1 (=> (smiles *X) (happy *X)))
```

then the assertion of

```
(smiles John)
```

in `sit1` will force PROSIT to assert the following in `sit1`, too:

```
(happy John)
```

When an expression, *expr*, is queried, PROSIT tries to evaluate the query, binding values to the variables in the query as the interpreter goes through the database. If this process fails at any stage, PROSIT backtracks to the previous stage in the search of a solution, and undoes all the bindings made along the incorrect path. The search will succeed in two cases:

1. *expr* unifies with an expression that is explicitly asserted in the current situation or its subsituations.

2. *expr* unifies with the *head* of a backward-chaining constraint (<= *head fact$_1$ fact$_2$ ...fact$_n$*) and finds a solution to all of *fact$_1$ fact$_2$ ...fact$_n$*, when queried in order.

PROSIT offers two types of unification. One is variable unification (V-unification), the other is parameter unification (P-unification). V-unification is the one familiar from Prolog and binds variables to objects. It occurs only in the query mode and its effects are undone when PROSIT backtracks. P-unification occurs only in the assertion mode. It is performed by explicitly stating that two parameters stand for the same object and can be unified. P-unification is one of the major differences between PROSIT and Prolog [12] in which atoms never unify.

PROSIT's querying mechanism is flexible. It is possible to use a variable in any part of a query, even in the predicate name or the entire query.

## 4.3 Applications

Although it offers a variety of constructs that can be used in inferencing for human-like reasoning, there have been few attempts to employ PROSIT in this style. One of the applications in which PROSIT was used is the treatment of identity. This aims to demonstrate the role of parameters in situation theory.

Parameters are means to keep track of the correspondence between the concepts in mind and real objects in the world, cf. Israel and Perry [24]. The idea can be exemplified by the discussion about Cicero. The famous Roman orator Cicero's first name is Tully. For someone who knows this identity, the answer to the question "Is Tully an orator?" would be yes. However, it is not possible to give the same answer for someone who is not aware of this identity.

In PROSIT, it is possible to express the difference between someone who knows the identity of Cicero and Tully, and someone who does not. PROSIT overcomes this identity problem by allowing assertions of situation-dependent equalities between parameters. This is done by P-unifying two objects.

If an individual is aware of the identity of Cicero and Tully, his knowledge will be classified by the situational parameter `sit1` which supports the following facts.

```
(!= sit1 (= cicero tully))
(!= sit1 (orator cicero))
```

Here, the former is a P-unification which states that Cicero and Tully are the same person; the latter states that Cicero is an orator.

On the other hand, the knowledge of someone who does not know this identity is classified by `sit2` where

```
(!= sit2 (orator cicero))
```

When asked the same questions in the two situations the system will respond

```
(!= sit1 (orator tully))
yes.
(!= sit2 (orator tully))
unknown.
```

A study on communication and inference through situations by Nakashima et al. [32], was the most serious attempt to make use of PROSIT. The study was mainly aimed to solve a problem that requires the cooperation of a group of agents in a multi-agent setting. Situation theory was used as a framework to represent common knowledge [3]. The idea behind this choice was to exploit the foundations of situation theory for analyzing information flow. Situation-theoretic principles were used to solve the "Three Wisemen Problem" [32] which will be covered in the next chapter.

## 4.4 PROSIT versus Situation Theory

The development of programming languages based on situation theory is a new trend, so it is worth examining how much PROSIT reflects situation-theoretic concepts and how much it deviates from them.

PROSIT represents infons as lists and this is similar to the representation of infons in situation theory. PROSIT has no special polarity argument in infons, but handles this feature using the predicate **no**. Thus, (*infon*) represents a positive infon whereas (**no** *infon*) stands for the negation of that infon. The only deficiency regarding infons in PROSIT appears in the notion of spatial and temporal locations. In PROSIT, it is possible to use location-indicating parameters in the argument places of relations, but this would be putting the individuals and locations in the same category. However, Devlin [17, p. 35] remarks that "...infons are built up out of entities called relations, individuals, locations, and polarities." Clearly, the majority of real-life facts pertain only to a certain region of space and a certain interval of time, and it is desirable to handle (spatial and temporal) locations.

PROSIT has situational parameters that are used to model abstract analogs of real situations. In that sense, they can be considered as abstract situations. They are associated with sets of infons. The definition of *supports* changes to:

> A situation $s$ *supports* an infon $i$, if $i$ is explicitly asserted to hold in $s$ or can be proved to hold by application of forward-chaining constraints in $s$.

As a result, *supports* reduces to simple set-membership and we can conclude that the situations in PROSIT are equivalent to abstract situations. PROSIT also supports the concept of constraints, but handles them in a different fashion. These come in three flavors in PROSIT: forward-chaining constraints, backward-chaining constraints, and forward- and backward-chaining constraints. (This classification is nowhere to be found in situation theory.) Built up on this classification, the designers of PROSIT came up with new definitions [33, p. 493]:

> An infon is *supported* by a situation if [it] is explicitly asserted to hold in the situation, or can be proved to hold by application of forward-chaining constraints in the situation.
> An infon is *permitted* by a situation if [it] is deduced through application of backward-chaining constraints.

It seems that this has no philosophical basis, but is offered because of implementation requirements. In fact, both methods (forward or backward) result in the same answers to queries. However, forward-chaining incurs a high cost at assertion-time, and backward-chaining incurs a high cost at query-time. Additionally, forward-chaining requires more computer memory. So what the expression "an infon is permitted in a situation" really means is that, the infon is supported by the situation but there is either no need or no space to store it. On the other hand, if implementation strategies are considered, it is a good thing to have such choices. It is left to the user to select what kind of constraints to use. For example, forward-chaining constraints can be used in applications where results may not be predictable, and backward-chaining can be used in diagnostic problems [51]. There are two additional points on which the constraints of PROSIT have been criticized (cf. Black [9, 10] and Tın [44]). The first point is that PROSIT's constraints are situated infon constraints, i.e., the constraints are about local facts within a situation rather than about situation-types. While this criticism seems to be valid, it is possible to simulate constraints that are not local to one situation (but are global). This can be achieved by introducing a situation which is global to all other situations and then asserting the constraint in this global situation. Because all other situations will be in this global situation, any constraint that is asserted here will apply to all situations. For example,

```
(!  (resp TopSit
      (<= (!= *Sit1 (touching *X *Y))
          (!= *Sit1 (kissing *X *Y)))))
```

states that if there is a situation in TopSit that supports a fact about "kissing," then that situation also supports a fact about "touching" on the same arguments.

The second criticism is that it is not possible to model conventional constraints in PROSIT. However, none of the existing systems is capable of performing this either.

Situation theory provides notions such as types and parameters. In PROSIT, some of these notions are hard to represent and some are not even possible. First of all, there is no typing in PROSIT. A variable can match any parameter or constant without due regard to types. In Chapter 2, we have defined $\dot{r}1$ as a restricted parameter ranging over all men kicking footballs. Once defined, $\dot{r}1$ will represent this subclass of individuals. But in PROSIT it is not possible to make this kind of parameter definitions that can be used throughout a program. The only thing one can try is to pose queries on restricted parameters. All men kicking footballs can be queried using the following expression:

```
(AND (kicking *a *b) (man *a) (football *b))
```

Although none of the variables above is restricted, the expression queries a restricted class of individuals.

PROSIT has no mechanism to define types either. As a consequence, there is a lack of situation-types. We cannot define a situation-type explicitly, i.e., there is no corresponding expression for defining all men kicking footballs. On the other hand, PROSIT can query a certain type of situation and put constraints between situation-types.

So the problem is that it is not possible to restrict a parameter or to assign a variable to a certain type. This also makes it impossible to define argument roles. Nevertheless, this deficiency does not prevent us from making queries about restricted parameters or enforcing constraints between situation-types.

Recalling the previous section, one may wonder why there are two different relations (owner and superchunk) doing very similar jobs. The major difference between these relations is not what the PROSIT manual [35, p. 15] says, i.e., that the predicate out will take the interpreter to the owner not to the superchunk. More importantly, the owner relation is defined between situations which are parent-child in the situation tree and the superchunk relation between two situations that are siblings in this tree.

The other two relations (subtype and subsituation) should also be examined carefully. At first glance, it seems that there is a similarity between these relations and the concept of inheritance in object-oriented programming [11]. However, in PROSIT the supersituation inherits all the infons from the subsituation, whereas in object-oriented programming it is the subclass that inherits the properties and methods from the superclass. Accordingly, it can be concluded that either the direction of inheritance is completely different in two paradigms or that the terms subsituation and subclass should not evoke object-oriented concepts.

Regarding the question of where one can use these relations, the example given in the PROSIT manual [35, p. 2] uses these relations to classify airplanes of type DC (DC-9,

DC-10, and so on). But from the situation-theoretic point of view, it is not correct to consider airplanes of type DC as a situation. An agent does not individuate DC type of airplanes as a situation and say, DC-9s as a subsituation of that situation. These can only be considered as a class and its subclass. This example surely suits well to object-oriented programming, but not to situation theory. Accordingly, PROSIT needs to draw a clear distinction between situations and classes.

One would be hard-pressed to find anything about inheritance, supersituations, and subsituations when one reads the essential documents on situation theory [8, 5, 17]. The only thing that seems related to these concepts is the "part-of" relation which is defined as follows [5, p. 185]:

> A situation $s_1$ is a part of a situation $s_2$ (denoted as $s_1 \trianglelefteq s_2$) just in case every basic state of affairs that is a fact of $s_1$ is also a fact of $s_2$.

However if $Sit_1 \trianglelefteq Sit_2$ is true, then the only comment we should make is that these two are defining the same situation, and that $Sit_2$ offers a more fine-grained description (using more infons than $Sit_1$).

# Chapter 5

# Situations and Epistemic Puzzles

## 5.1 Epistemic Puzzles

Epistemic puzzles deal with agents and their knowledge. This can be either in the form of individual knowledge or common knowledge (mutual information) in a multi-agent setting. The ontology of these puzzles include the *agents* whose knowledge we try to represent, $A = \{a, b, c, \ldots\}$, the *knowledge* each agent has, $K = \{K_a, K_b, K_c, \ldots\}$, and the *facts* mentioned in the statement of the puzzle. If we let all the facts in a puzzle make up the set $F = \{f_1, f_2, f_3, \ldots\}$ (where each $f_i$ is a relation that holds among the agents and objects that exist in the puzzle), then each $K_i$ is a subset of $F$. The primary question to be answered in these epistemic puzzles is generally about the facts that the agents are aware of. So a puzzle might ask if an agent, say $x$, is aware of the fact $f_i$, i.e., whether $f_i \in K_x$ is true. However, this representation fails to handle two main issues of knowledge[3, 6]: the circularity of knowledge (i.e., if $a$ knows $f_3$, then he knows that he knows $f_3$, ad infinitum) and deductive omniscience (i.e., if $a$ knows that $p$ and $p$ entails $q$, then $a$ knows that $q$). For this representation to handle circularity of knowledge it should be extended such that each $K_i$ is an element of itself. So if $a$ knows the facts $f_1$, $f_3$, and $f_4$, then $K_a = \{f_1, f_3, f_4, K_a\}$. To achieve deductive omniscience the definition of the *facts* should be extended. In addition to simple relations that hold among agents, rules of the form "if ... then ..." should also be considered as *facts*.

To elucidate the definitions above, we show how they can be used to represent common knowledge. In a card game, John has the ace of spades and Mary has the queen of spades. Jack comes and looking at her cards announces that Mary has the queen of spades. At this point, each agent's knowledge is represented as follows:

$K_{john} = \{$*(has john ace-of-spades)*, $K_{john}, K_{common}\}$
$K_{mary} = \{K_{mary}, K_{common}\}$
$K_{common} = \{$*(has mary queen-of-spades)*, $K_{common}\}$

We now describe, via an example, what an epistemic puzzle looks like [34]:

> Two logicians place cards on their foreheads so that what is written on the card is visible only to the other logician. Consecutive positive integers have been written on the cards. The following conversation ensues:
> A: "I don't know my number."

> B: "I don't know my number."
> A: "I don't know my number."
> B: "I don't know my number."
> ... $n$ statements of ignorance later ...
> A or B: "I know my number."

What is on the card and how does the logician know it?

Note that the facts that we are after are restricted. We are only interested in the numbers on the cards on the foreheads, not in the colors or shapes of the cards. Here, both logicians know some facts, and as the conversation proceeds they generate new facts. At the end, one of them finds out what the number on his forehead is. The aim of this study is to simulate the way the agent holds information about the situation he happens to be in and the way he reasons about this information.

There have been many attempts in AI to deal with knowledge and information, and the most common tool used in tackling the fundamental problems posed by these concepts was classical logic (predicate logic) or extensions of it such as modal, temporal, and deontic logics [2, 23, 27]. All these attempts were of a strictly mathematical nature, and therefore all were within the existing pure mathematics paradigm [5]. On the other hand, situation theory emerged as a realistic theory of information. First, an empirical study of information was made [17]. This was followed by both the application of the existing mathematical techniques and the development of new mathematical tools. In that respect, situation theory is tailor-made for problems involving knowledge and information.

In the following section, we will compare how the classical approach and the situated approach handle epistemic puzzles.

## 5.2 Previous Approaches

In this section we will examine three different approaches used in solving epistemic puzzles. First we will analyze how Smullyan solves his famous knights-and-knaves puzzles using symbolic logic. In [42], Smullyan introduces a number of puzzles about liars and truth-tellers to warm up the layman. Most of the events in the puzzles take place on an island, viz., the Island of Knights and Knaves. On this imaginary island the following three propositions hold:

1. Knights always make true statements.

2. Knaves always make false statements.

3. Every inhabitant is either a knight or a knave.

The aim of the puzzles is to decide whether an inhabitant is a knight or a knave using the statements he makes. Assume that $P$ is a native of the Island of Knights and Knaves. Let $k$ be the proposition that "$P$ is a knight." Suppose $P$ utters a proposition $X$. In Smullyan's puzzles the reasoner knows neither the truth value of $k$ nor the truth value of $X$, i.e., he does not know whether the native is a knight or knave, and he does not know if the asserted proposition is true or false. The only thing he knows is that $P$ is a knight if and only if $X$ is true. So he knows that the proposition $k \equiv X$ is true. So the sentence "$P$ asserts $X$" is translated as $k \equiv X$. We will show how this fact helps in the solution of the problem where the native $P_1$ states that he and his wife, $P_2$, are both knaves. Now, $k_1$ is the proposition that $P_1$ is a knight, and $\neg k_1$ that he is a knave. Similarly $\neg k_2$ is the proposition that $P_2$ is a knave. Translating to symbolic logic, the reasoner knows that $k_1 \equiv (\neg k_1 \wedge \neg k_2)$. At this point, the domain of the problem changes from knowledge to symbolic logic: Given two propositions $k_1$ and $k_2$ such that $k_1 \equiv (\neg k_1 \wedge \neg k_2)$, what are the truth values of $k_1$ and $k_2$? Using a truth table one can easily verify that the only case in which $k_1 \equiv (\neg k_1 \wedge \neg k_2)$ is when $k_1$ is false and $k_2$ is true.

Although there is a very interesting translation here from the domain of knowledge to the domain of symbolic logic, the question "Is this the way an intelligent agent handles such problems?" should be carefully considered. Would an intelligent agent use a truth table to decide who is lying and who is telling the truth?

In the solutions of some other epistemic puzzles, rather than explaining the way an agent reasons throughout the puzzle, it is proven that the final result that the agent has reached is correct. For example, in the puzzle about cheating husbands this is done using induction [34]:

> The queen of the matriarchal city-state of Mamajorca, on the continent of Atlantis, have a long record of opposing and actively fighting the male infidelity problem. Ever since technologically-primitive days of Queen Henrietta I, women in Mamajorca have been required to be in perfect health and pass an extensive logic and puzzle-solving exam before being allowed to take a husband. The queens of Mamajorca, however, were not to show such competence.
>
> It has always been common knowledge among the women of Mamajorca that their queens are truthful and that the women are obedient to the queens. It was also common knowledge that all women hear every shot fired in Mamajorca. Queen Henrietta I awoke one morning with a firm resolution to do away with the infidelity problem in Mamajorca. She summoned all of the women heads-of-households to the town square and read them the following statement:
>
> "There are (one or more) unfaithful husbands in our community. Although none of you knew before this gathering whether your own husband was faithful, each of you knows which of the other husbands are unfaithful. I forbid you to discuss the matter of your husbands fidelity with anyone. However, should you discover your husband is unfaithful, you must shoot him on the midnight of the day you find about it."
>
> Thirty nine silent nights went by, and on the fortieth night, shots were heard. How did the wives decide on the infidelity of their husbands?

As a solution to this problem, the theorem stating that if there are $n$ unfaithful husbands they will be shot on the midnight of the $n^{th}$ day, is proven. For $n = 1$ there would be one unfaithful husband. His wife would immediately realize that he is the unfaithful one, just after hearing the queen's statement, because she definitely knows that there is no other unfaithful husband. Assume the claim holds for $n = k$, i.e., if there are $k$ unfaithful husbands they would be shot on the $k^{th}$ night. It could be proven that the claim also holds if there are $k + 1$ unfaithful husbands. In that case, every cheated wife would know $k$ unfaithful husbands. As all the cheated wives are logically competent, they know that if there are $k$ unfaithful husbands then those husbands will be shot on the $k^{th}$ night. As none of the cheated wives can prove that their husband is unfaithful, no shots are fired during the first $k$ nights. Because no shots are fired on the $k^{th}$ night, the cheated wives decide that there are more than $k$ unfaithful husbands and that their own husband is unfaithful too. So the unfaithful husbands are shot on the $k+1^{st}$ night.

However, rather than explicating how the cheated wives decide that their husbands are unfaithful, this proof demonstrates that their decision is correct.

The third approach used in solving these puzzles is the most realistic one. It explains how the agents in these puzzles reason about the situations they find themselves in. A slight blemish of this approach is that, it is informal. For example, the solution of the puzzle where the native $P_1$ states that he and his wife, $P_2$, are both knaves is given as follows [42, p. 16]:

> If the husband were a knight, he would never have claimed that he and his wife were both knaves. Therefore he must be a knave. Since he is a knave, his statement is false; so they are not both knaves. This means his wife must be knight. Therefore he is a knave and she is a knight.

This informal solution seems to be the right way to handle these puzzles. What we will try to do in the sequel is in some sense to formalize this using situation-theoretic concepts.

## 5.3 The Situated Approach

Situation theory, as mentioned earlier, is tailor-made for the problems involving knowledge and information. It provides a group of features that motivated the design of the language PROSIT which is especially suitable for writing commonsense reasoning programs.

One of these features is that "situations are first-class citizens of the theory." This feature combines reasoning *in* situations and *about* situations. More specifically, situations can be arguments to relations. Therefore situation theory should not only be considered as a theory of relations *in* situations, but also of relations *among* situations.

Both individual and common knowledge are represented as situations. These situations consist of a number of infons representing the facts an agent is aware of. So if Jack knows that John has the ace of spades and that Mary has the queen of spades, then the situation representing Jack's individual knowledge, i.e., `jack_knows`, will consist of two infons (discarding any unrelated stuff):

```
(!= jack_knows (has john ace_of_spades))
(!= jack_knows (has mary queen_of_spades))
```

This representation is analogous to the definition of the the agents' knowledge in epistemic puzzles. For example, Max's knowledge of the facts $fact_1$, $fact_2$, and $fact_3$, i.e., $K_m = \{fact_1, fact_2, fact_3\}$, is represented as `(!= max_knows (infon_1) (infon_2) (infon_3))` where `max_knows` stands for the individual knowledge of Max and `infon_i` for the the infon stating $fact_i$.

An important advantage of using situations to represent knowledge is that it is possible to express some statements that are not expressible in logic. For example, the statement

> "I know a man who drinks wine every night."

can be most closely rendered in predicate logic by the following expression:

$$(\exists x)[know(I, x) \land drinks\_wine(x)]$$

However, this expression can also be interpreted as: "I know a man, and that man drinks wine every night. (I don't know whether he drinks wine every night.)"[1] Using situations to represent knowledge we would use the following infons to express the statement.

```
(!= i_know (man *x))
(!= i_know (drinks_wine *x))
```

On the other hand, if I didn't know whether he drinks wine every night, then the second infon would not hold. (It would hold in that agent's individual knowledge who is aware of that fact.)

Another feature of situation theory that helps formalize epistemic concepts is the general treatment of partial information. As mentioned previously, situations are partial, i.e., they do not define the truth or falsity of all relations on all objects in the domain. Assume that there are two agents, Mary and John, facing each other stand in a room, and there is a cat behind Mary. As John is seeing the cat, the situation that models his knowledge will support the fact that there is a cat in the room:

```
(!= john_knows (in_room cat))
```

However, in the situation representing Mary's knowledge there should not be an infon about the cat being in the room. When a query is made about the cat in the situation representing Mary's knowledge, the answer should not be "no" but rather "unknown".

Additionally deductive omniscience and circularity of knowledge are handled elegantly. Logical omniscience is supported by the feature of "informational constraints" in situation theory. Constraints are the main tool for information flow. They are relations that hold between two situation-types, therefore are also considered as infons. For example, if Jack is aware of (or attuned to) the constraint that everything that smiles is happy, and knows that John is smiling, then he deduces the fact that John is happy. The constraint is represented as follows:

```
(resp jack_knows (=> (smiles *X) (happy *X)))
```

The circularity of knowledge is modeled via "self-referential expressions" which is another important feature of situation theory. Situations are members par excellence of the ontology of situation theory; therefore, they can be used as constituents of infons. This property makes it possible to define circularity. For example, if common stands for the situation holding the facts that are common to every agent, i.e., the common knowledge situation, and jack_knows stands for Jack's individual knowledge, then the following expressions state that Jack knows everything that is common knowledge:

```
(!= jack_knows common) or
([_ common jack_knows)
```

---

[1] This is very similar to the argument Barwise advances about representing perception [4, p. 21].

In PROSIT the second representation expresses the subchunk relation. It can be translated as "common is totally described by the infons in jack_knows." So if (infon) holds in common, then (!= common (infon)) holds in jack_knows.

Using the subchunk relation it is straightforward to define circularity. ([_ jack_knows jack_knows) will generate a self-referential situation, and as jack_knows stands for the individual knowledge of Jack, it will provide the circularity of Jack's individual knowledge. So if Jack knows that Mary has the queen of spades, then all of the following infons will hold:

```
(!= jack_knows (has mary queen-of-spades))
(!= jack_knows (!= jack_knows (has mary queen-of-spades)))
(!= jack_knows (!= jack_knows (!= jack_knows (has mary queen-of-spades))))
```

Jack knows that he knows that Mary has the queen of spades, he knows that he knows that he knows that Mary has the queen of spades, and so on.

The final feature of situation theory that led us to using it as a framework for epistemic puzzles is the "situatedness of information and constraints" [31, 43]. Each infon or constraint exists in a situation (more formally, is supported by a situation). Consequently, each infon or constraint has an interpretation according to the situation it exists in. This can be considered as "context dependence" [1].

To clarify the argument above, consider a constraint that deduces facts about the height of individuals. Let both Mike and John be 185 cm tall. Both are aware of the fact that if someone is higher than 185 cm, then that individual is taller than Mike and John. The following represents the constraint that is supported by Mike's (or John's) knowledge:

```
(resp mike_knows (<= (taller *y *x)
                     (me *x)
                     (height *y *h)
                     (> *h 185)))
```

If John knows that Bob is 175 cm tall and Mike knows that Bill is 195 cm tall, then Mike will deduce the fact that Bill is taller than himself, viz.

```
(!= mike_knows (taller bill mike))
```

but John will not be able to deduce anything using the previous constraint.

The same argument holds for infons. Assume a case which Holmes and Watson are working on. Consider a theft in which the door of the flat that the thief broke into is not fractured and that the windows are closed. Although both Holmes and Watson are aware of these facts, only Holmes is able to deduce the fact that the thief had the key to the door:

```
(!= holmes_knows (no (broken door)) (closed windows))
(!= watson_knows (no (broken door)) (closed windows))
```

This is because only Holmes finds out that the thief has a key, using the following constraint:

```
(resp holmes_knows (=> (and (no (broken door))
                            (closed windows)
                       (has thief key))))
```

This example demonstrates that the same infon can generate different facts in different contexts; a system should simulate this capability if it is trying to perform human-like reasoning.

From the argument above, it can be concluded that the main advantage of using situation theory in representing knowledge is the conceptual clarity and elegance it offers. Epistemic puzzles can be modeled without much effort. All the tools required for such a modeling are already present in the domain of situation theory. This, as mentioned numerous times throughout this paper, is due to the fact that situation theory is a natural theory of information.

### 5.3.1 The Three Wisemen Problem

The solution of the "Three Wisemen Problem" [32] in PROSIT is, to our best knowledge, the only serious attempt to use situation-theoretic constructs in the resolution of epistemic puzzles. The main aim is to show how to use common knowledge computationally in solving problems involving cooperation of multiple agents. It turns out that the situation-theoretic aspects of PROSIT (reasoning *about* situations and *in* situations) generate an intuitive and simple solution for this hypothetical problem [32, p. 79]:

> Three wisemen are sitting at a table, facing each other, each with a white hat on his head. Someone tells them that each of them has a white or red hat but that there is at least one white hat. Each wiseman can see the others' hats but not his own. If a fourth person asks them whether they know their own color, then the first two wisemen will answer no, but, after that, the third one will answer yes.

The available facts in the problem can be categorized into two groups: facts that all wisemen are aware of, and facts that are known individually. Facts such as that there are three agents $A$, $B$, and $C$, that all agents are wise, and that each agent is wearing either a white or a red hat are known by all three wisemen. On the other hand, the fact that say, $B$ and $C$ are wearing white hats is known only by $A$.

There are two ways for an agent to decide that his hat is white. The first is when the other two wisemen have red hats. The second is when his assumption of having a red hat causes a contradiction. The approach followed by [32] is to use the latter in order to solve this problem. $A$ assumes that he has a red hat. After $B$ and $C$ answers no, $A$ concludes that $C$ should have said yes (because from $B$'s answer $C$ concludes that at least one of $A$ and $C$ is wearing a white hat) if he were wearing a red hat. So he knows that he is wearing a white hat. PROSIT's tree hierarchy of situations makes it rather easy to represent this (Figure 5.1).

The program that models this puzzle has a deficiency. It is not possible to distinguish the following two cases that would result in different ways:

- The logicians answering the questions in sequence.

- The logicians answering all at the same time (in which case none of them can decide on their color).

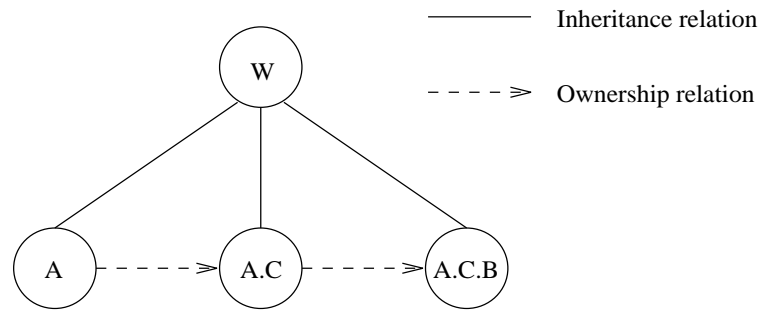

Figure 5.1: The Three Wisemen Problem. The facts known to all wisemen are kept in situation W. The facts that A knows are kept in situation A. The facts that A knows that C knows are kept in situation A.C. The facts that A knows that C knows that B knows are kept in situation A.C.B.

### 5.3.2 Smullyan's Puzzles

These puzzles are epistemic in the sense that knights 'reflect' their individual knowledge and beliefs while knaves 'reflect' the contrary of them. A simple puzzle of this type is the following [42, pp. 15–16]:

> The census-taker Mr. McGregor once did some fieldwork on the Island of Knights and Knaves. On this island, women are also called knights and knaves. McGregor decided on this visit to interview married couples only. McGregor knocked on one door; the husband partly opened it and asked McGregor his business. "I am a census-taker," replied McGregor, "and I need information about you and your wife. Which, if either, is a knight, and which, if either, is a knave?"
> "We are both knaves!" said the husband angrily as he slammed the door. What type is the husband and what type is the wife?

The solution is as follows [42, p. 16]:

> If the husband were a knight, he would never have claimed that he and his wife were both knaves. Therefore he must be a knave. Since he is a knave, his statement is false; so they are not both knaves. This means his wife must be knight. Therefore he is a knave and she is a knight.

As it can be seen from the solution of the puzzle, when a reasoner is asked to solve this puzzle he first makes assumptions. Then based on these assumptions he considers a hypothetical world and tries to find out if there are any incoherencies in this hypothetical world. If an incoherency exists he concludes that his assumption is wrong and totally forgets about that hypothetical world. The reasoner continues to make new assumptions (while learning something from the previous failures) until he finds all the solutions of the puzzle, i.e., the coherent hypothetical worlds (Figure 5.2). In the puzzle above, first it was assumed that the husband is a knight, but this assumption led to failure because a knight can never claim that he is a knave (an incoherency). So it was decided that the husband is a knave.

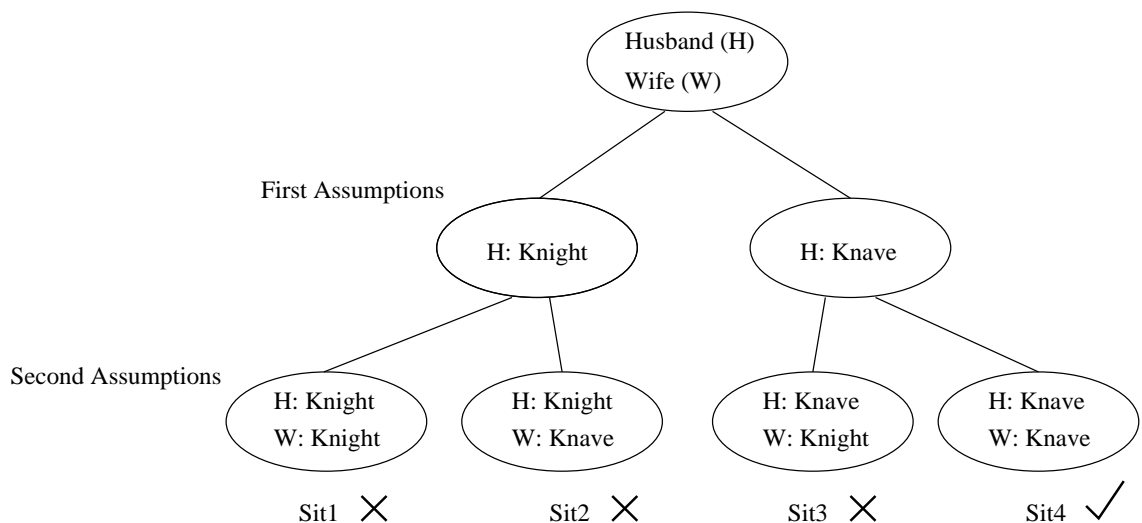

Figure 5.2: The hypothetical worlds created by the reasoner for the census-taker problem. There is only one world (Sit4) coherent with the statement the husband uttered.

Examining the structure of these puzzles one will notice properties that are suitable for a situation-theoretic representation:

- Actions always take place in a clearly defined context, i.e., the Island of Knights and Knaves.

- There are abstract individuals, properties, and relations (e.g., being a knight, being on the island, and so on).

- There are well-defined rules that invariably hold on the island (e.g., knights always make true statements).

As mentioned previously, a system to solve these puzzles should be able to make human-like reasoning. There are three main properties that enable PROSIT to simulate human-like reasoning. The first one is situated programming, i.e., infons and constraints are local to situations. The second is PROSIT's situation tree structure, with which one can represent nested knowledge/belief (e.g., "$A$ thinks that $B$ believes that $C$ knows ..."). The third is the use of incoherency to generate new information. Now, it is time to see how PROSIT solves these puzzles. The following puzzle [42, pp. 23–24] will be exploited to explain our approach:

> This is the story of a philosopher—a logician, in fact—who visited the cluster of islands and fell in love with a bird-girl named Oona. They were married. His marriage was a happy one, except that his wife was too flighty! For example, he would come home late at night for dinner, but if it was a particularly lovely evening, Oona would have flown off to another island. So he would have to paddle around in his canoe from one island to another until he found Oona and brought her home. [...] On one occasion, the husband came to an island in search of Oona and met two natives $A$ and $B$. He asked them whether Oona had landed on the island. He got the following responses:

```
; testing the coherency of a situation requires a
; translation of the uttered sentences to what they
; really mean
(!  (resp island (<= (coherent)
      (means P1 *sentence *translation)
      (means P2 *sentence2 *translation2)
      (and *translation *translation2))))
; every sentence uttered by a knight is true
(!  (resp island (<= (means *x *sentence *sentence)
      (says *x *sentence)
      (knight *x))))
; any sentence uttered by a knave is false
(!  (resp island (<= (means *x *sentence (no *sentence))
      (says *x *sentence)
      (knave *x))))
```

Figure 5.3: Three main constraints of the puzzle about Oona.

$A$: $B$ is a knight, and Oona is on this island.
$B$: $A$ is a knave, and Oona is on this island.
Is Oona on this island?

The solution of this puzzle will make use of various properties of PROSIT, including inheritance. As the solution is based on creating hypothetical situations and testing their coherency, it is useful to have a situation, say *island*, from which all the hypothetical situations will inherit some essential facts that will not change from one situation to another. For example, the fact that the native $A$ says "$B$ is a knight, and Oona is on this island" will hold in every hypothetical situation. Therefore this fact is kept in *island*. Similarly, the rules stating that knights always make true statements and that knaves always make false statements are kept in *island*. The three main constraints used in the solution of this puzzle are shown in Figure 5.3.

The first step of the solution, i.e., making assumptions about the natives, is simulated by creating hypothetical situations. Each hypothetical situation represents a different combination of assumptions. A reasoner can assume the native $A$ to be a knight or a knave, the native $B$ to be a knight or a knave, and Oona to be on the island or not. So, the program will generate eight ($2^3$) hypothetical situations. The following are two hypothetical situations (Sit1, Sit2) that we will be examining throughout this section:

```
Sit1:   (knight A) (knave B) (on_island Oona)
Sit2:   (knave A) (knave B) (not_on_island Oona)
```

The next step is to generate the infons that hold in the hypothetical situations. If a knight makes a statement, it means that this statement holds in that situation. On the other hand, if a statement is made by a knave, it is concluded that the negation of that statement holds in the situation. So the following infons hold in the hypothetical situations Sit1 and Sit2:

```
Sit1: (and (knight B) (on_island Oona))
      (no (and (knave A) (on_island Oona)))
```

```
    Sit2: (no (and (knight B) (on_island Oona))
          (no (and (knave A) (on_island Oona)))
```

The final step is to check the hypothetical situations and to discard the ones that are incoherent. The coherent situations are then the solutions of the puzzle. In the previous case `Sit1` is one of the incoherent hypothetical situations to be discarded, and `Sit2` is a solution (in fact, the only solution):

    Sit1: (and (knight B) (on_island Oona)) (from the second step)
          (knave B) (from the first step)
          Incoherent!
    Sit2: Coherent, therefore A and B are knaves and Oona is not on the island.

Smullyan's solution is as follows [42, p. 26]:

> A couldn't possibly be knight, for if he were, then B would be a knight (as A said), which would make A a knave (as B said). Therefore A is definitely a knave. If Oona is on the island we get the following contradiction: It is then true that A is a knave and Oona is on the island, hence B made a true statement, which makes B a knight. But then A made a true statement in claiming that B is a knight and Oona is on the island, contrary to the fact that A is a knave! The only way out of the contradiction is that Oona is *not* on the island. So Oona is not on this island (and, of course, A and B are both knaves).

The simpler puzzle given earlier, i.e., the one about Mr. McGregor, is solved in a similar fashion. There are two natives, $H$ and $W$, in the puzzle. Each can be either a knight or a knave. So there will be four hypothetical situations (Figure 5.2):

    Sit1:   (knight H) (knight W)
    Sit2:   (knight H) (knave W)
    Sit3:   (knave H) (knight W)
    Sit4:   (knave H) (knave W)

After the generation of new infons using the statement uttered by $H$, the hypothetical situations will consist of the following:

    Sit1:   (knight H) (knight W) (and (knave H) (knave W))
    Sit2:   (knight H) (knave W) (and (knave H) (knave W))
    Sit3:   (knave H) (knight W) (no (and (knave H) (knave W)))
    Sit4:   (knave H) (knave W) (no (and (knave H) (knave W)))

Among these hypothetical situations the only coherent one is `Sit3`, which states that $H$ is a knave and $W$ is a knight.

It is time to examine how PROSIT finds out about these incoherencies. As it is seen from the examples above a distinguishing feature of PROSIT is that it allows incoherency in situations. A situation may support both $i$ and (`no` $i$). This should not be considered as a contradiction in the system, but merely a contradiction in the situation, which means that the situation is incoherent (cannot be actual). This kind of incoherency can

```
; if a native is a knight, he definitely is not a knave
(!  (resp island (=> (knight *x)
         (no (knave *x)))))
; if a native is a knave, he definitely is not a knight
(!  (resp island (=> (knave *x)
         (no (knight *x)))))
; (no (and *st1 *st2)) is equivalent to
; (or (no *st1) (no *st2))
(!  (resp island (<= (means *x (or (no *st1) (no *st2)))
         (says *x (and *st1 *st2))
         (knave *x))))
; (no (or *st1 *st2)) is equivalent to
; (and (no *st1) (no *st2))
(!  (resp island (<= (means *x (and (no *st1) (no *st2)))
         (says *x (or *st1 *st2))
         (knave *x))))
```

Figure 5.4: The constraints about negative knowledge.

be adequately used to get new information. In the example above, there is a situation (Sit1) that supports both (knight H) and (knave H). (knave H) is equivalent to (no (knight H)) (using the rules in Figure 5.4), therefore both (knight H) and its negation are supported by the situation. The situation is incoherent and the assumptions have failed. One final comment on PROSIT is that it does not apply the predicate no over the predicates and and or, therefore two additional constraints should be explicitly defined in order to achieve this (Figure 5.4).

### 5.3.3 The Cheating Husbands Puzzle

The cheating husbands puzzle, studied in Section 5.2, is well-known from folklore [19] and has long been the primary example to illustrate the subtle relationship between knowledge, communication, and action in a distributed environment [16]. The puzzle involves an initial step in which a set of facts is announced publicly, thereby becoming common knowledge.

Moses et al. [30], using a number of variants of the puzzle, try to describe what happens when

1. synchronous communication,
2. asynchronous communication, and
3. ring-based communication

channels are used to communicate the protocol to be followed, i.e., announce the orders of the queen. The distributed computational point of view is mainly interested in the types of protocols, the delays and bounds in communication, and whether the communication is fault-tolerant or not. For example, instead of making an announcement at the town-square, the queen sends letters to all wives which makes the communication asynchronous. Similarly, to test whether the system is fault-tolerant, another version of the puzzle in which wives are disobedient, i.e., wives that talk to each other about their husbands, is used.

On the other hand, we are interested in the way agents reason about knowledge, assuming that communication is totally synchronous and reliable. We are using the puzzle to illustrate how intelligent agents reason in a multi-agent system, and how they represent each others' knowledge.

The tree wisemen problem is a special case of this puzzle where the number of agents is restricted to three. In this puzzle all the wives in Mamajorca know each other. They know that a husband is either faithful or unfaithful. On the other hand, none of the wives know whether their husbands are faithful or not. Because all of these facts are common to all the wives in Mamajorca, they are supported by the situation `wives` which holds the infons that are common to all individuals in the puzzle. Some of the relations require an argument that indicates the temporal location. The temporal location is represented by an integer, $n$, which indicates the $n$th night after the queen has made the announcement. Every silent night after the announcement is regarded as the wives not being able to decide about their husbands fidelity. In modeling this puzzle, we are only interested in what the the wives in Mamajorca know about the fidelity of the husbands in Mamajorca.

We now analyze the case where there are three unfaithful husbands. After the second silent night following the announcement the queen made, `b`'s (a wife) not knowing whether her husband is faithful or unfaithful is represented as

```
(!= wives (no (!= b (faithful b 2))))
(!= wives (no (!= b (unfaithful b 2))))
```

Let `a` be one of the wives whose husband is unfaithful. Throughout the two silent nights she knows who the other cheated wives are (say, `b` and `c`).

```
(!= a (unfaithful b 1))
(!= a (unfaithful c 1))
(!= a (unfaithful b 2))
(!= a (unfaithful c 2))
```

A wife whose husband is unfaithful, would realize this fact either if none of the other wives are cheated (because the queen declared that there are some unfaithful husbands) or if her assumption that her husband is faithful generates a contradiction. The latter can be considered as proof-by-contradiction.

The way a wife decides that her husband is unfaithful is via making assumptions and checking whether an assumption causes any incoherencies (Figure 5.5). Let `a` be the wife who is reasoning. In the constraint that models the way a wife would reason in such a situation, the premise `(me *x)` would bind `*x` to the situation this constraint is activated in, e.g., `a`. The next two premises bind the variables `*y` and `*z` to the other cheated wives `b` and `c`. The premises

```
(!  ([_ wives *y))
(!  (@< wives *y))
```

indicate that `b`, i.e., the individual bound to the variable `*y`, knows that the facts supported by `wives` are common to all wives (subchunk relation), and is aware of all the facts that are

```
; A wife knows that her husband is unfaithful if the
; assumption that her husband is faithful results in
; an incoherent situation.
(!  (resp wives (<= (unfaithful *x *time)
             (me *x)
             (wife *y)
             (wife *z)
             (not (= *x *y))
             (not (= *z *x))
             (not (= *z *y))
             (!  ([_ wives *y))
             (!  (@< wives *y))
             (bind-lisp *pre (- *time 1))
             (!  (!= *y (faithful *x *pre)))
             (transfer_knowledge_about_third *y *z)
             (incoherent *y))))
```

Figure 5.5: The constraint that decides the fidelity of a husband by making assumptions and searching for incoherencies.

supported by `wives` (subtype relation). Next `a` assumes that her husband is faithful. She knows that if her husband were faithful, the other wives would know it. In the program, this assumption is made by asserting the fact that `a`'s husband is faithful in the situation that holds the facts that `a` knows that `b` knows, via the premise

```
(!  (!= *y (faithful *x *pre)))
```

The variable `*pre` is assigned to the value `*time`−1 (using the `bind-lisp` predicate that makes use of Lisp functions), where `*time` indicates the night on which the reasoning is made. Moreover, the facts that `a` knows about `c`'s husband are also asserted into the situation supporting the facts that `a` knows that `b` knows, because what `a` knows about `c`'s husband, `b` knows it too. This is achieved by the constraint `transfer_knowledge_about_third`. The final step is to check if the assumption `a` made would cause any incoherency. This is realized by the constraint `incoherent` which checks if a situation supports a fact we know it does not support. It should be noted that this rule implicitly expresses the fact that if someone is not faithful, he is unfaithful.

The constraint that transfers knowledge about the third individual (Figure 5.6) is a good example of the use of the situation tree hierarchy. If `a` knows on the second night after the announcement was made that `b`'s husband is unfaithful then she knows that `c` knows it too.

```
(!= a (unfaithful b 2))
```

```
(!= a (!= c (unfaithful b 2)))
```

So an infon supported by the situation `a` is copied to another situation `a.c` using the procedure `transfer_knowledge_about_third` .

```
; A hypothetical situation is incoherent if it
; supports a fact we know it does not support.
(!  (resp wives (<= (incoherent *y)
                    (no (!= *y *x))
                    (!= *y *x))))

; If the wives *x and *y know the character of the
; third wife's (*z) husband, they know that each of them
; knows it.  So if (!= *x (character *z)), then
; (!= *x (!= *y (character *z))) should be asserted.
(!  (resp wives (<= (transfer_knowledge_about_third *y *z)
                 (or
                  (and
                   (character *character)
                   (*character *z *time)
                   (bind-lisp *pre (- *time 1))
                   (!  (!= *y (*character *z *pre)))
                   )
                  (true)))))
```

Figure 5.6: Constraints that are used to find incoherencies and to transfer knowledge about the third party.

The constraint `incoherent` (Figure 5.6) checks whether a situation is coherent or not. This is achieved by searching for an infon that is supported by that situation with both positive and negative polarities.

To clarify the explanations made above, consider how a would reason until she finds out that her husband is unfaithful. a knows that b and c are being cheated:

```
(!= a (unfaithful b 3))
(!= a (unfaithful c 3))
```

a wishes to learn whether her husband is faithful or not. She assumes that her husband is faithful. She knows that if her assumption were true, then b would be aware of this fact. She also knows that b knows the fact that c is being cheated.

```
(!= a (!= b (faithful a 2)))     ; a's assumption
(!= a (!= b (unfaithful c 2)))   ; transferred knowledge about c
```

b did not shoot her husband on the second night, i.e., she did not know that her husband was unfaithful. So, an incoherency would occur, if she shot her husband on the second night, i.e., if she knew that her husband was unfaithful. (This would make a's assumption false, and mean that a's husband is unfaithful.) To decide on the truth of her assumption, a should learn whether b could have decided that her husband is faithful or unfaithful. b could decide about her husbands fidelity, just like a did. b would also assume that her husband was faithful. Then b would know that c would also know this fact on the first night after the announcement was made. `b` would also know that `c` would have known that `a`'s husband is faithful.

```
(!= a (!= b (!= c (faithful b 1)))) ; b's assumption
(!= a (!= b (!= c (faithful a 1)))) ; transferred knowledge about a
```

However this assumption of `b` would lead to a contradiction, because if `c` had known that both `a`'s and `b`'s husbands are faithful, then she would have immediately decided that her husband is unfaithful and shoot him.

Because of this incoherency, `b` must decide that her husband is unfaithful, on the second night, and shoot him. In other words, if `a`'s assumption about her husband were true then `b` would have shot her husband on the second night. But this did not happen, which means that `a`'s assumption that her husband is faithful fails. `a`'s husband is unfaithful and she shoots her husband on the third night after the announcement was made. Making the same reasoning `b` and `c` also shoot their husbands.

### 5.3.4 The Facing Logicians Puzzle

The facing logicians puzzle is another famous puzzle which can be considered to be epistemic. The puzzle statement was given in Section 5.1. Assume that the first logician, $A$, has the number 4 on the card on his forehead, and the other logician, $B$, has the number 3.

$A$ knows that the number on the forehead of $B$ is 3, while the $B$ knows that $A$ has the number 4 on his forehead. It is common knowledge to both of the logicians that the numbers on their foreheads are positive. (Both of the logicians are aware of the fact that common knowledge *is* common.)

```
(!= a (num b 3))
(!= b (num a 4))
(!= common (no (num a 0)))
(!= common (no (num b 0)))
([_ common a)
(@< common a)
([_ common b)
(@< common b)
```

Facts that are common knowledge are known by all the individuals and it is known that these facts are common (Figure 5.7). The subchunk relation, `[_`, is used to indicate that the individual knows that the facts supported by the situation `common` are common. The subtype relation, `@<`, on the other, indicates that any infon that is supported by the situation `common` is also supported by the situation representing the individual's knowledge.

It is also common knowledge that the numbers are consecutive. So if a logician knows that the number on the forehead of the other logician is $n$ and if he also knows that the number on his own forehead is not $n-1$ then he definitely knows that the number on his forehead is $n+1$.

Assume that $B$ is the one who is asked if he knows what the number on his forehead is. $B$ would answer "no" because he does not have enough knowledge to make a decision. He could only answer "yes" if the number on the forehead of $A$ were 1. Then he could easily

```
       ; if the number on the other logician's forehead is n and
       ; if the logician knows that the number on his forehead
       ; is not n-1, then the number on his forehead is n+1
       (!  (resp common (<= (know *x)
                  (me *x)
                  (logician *y)
                  (not (= *x *y))
                  (num *y *z)
                  (bind-lisp *a (- *z 1))
                  (no (num *x *a))
                  (bind-lisp *k (+ *z 1))
                  (!  (num *x *k)))))
```

Figure 5.7: The constraint with which a logician finds out the number on his forehead.

```
       (!  (resp common (<= (no (know *x))
                  (me *x)
                  (logician *y)
                  (not (= *x *y))
                  (not (!= common (num *y *k)))
                  (no (num *y *z))
                  (bind-lisp *a (+ *z 1))
                  (!  (!= *y (num *x *a)))
                  (!  ([_ common *y))
                  (!  (@< common *y))
                  (incoherent *y)
                  (clear *y)
                  (not (num *x *s))
                  (!  (!= common (no (num *x *a))))))))
```

Figure 5.8: The constraint that generates the numbers that cannot be on the forehead of a logician.

deduce the fact that the number on his forehead was 2. $B$'s answer, however, will make $A$ to learn that the number on his ($A$'s) forehead is not 1.

How does $A$ come to such a decision? Well, he makes assumptions about the number on his forehead. He assumes that the number on his forehead is 1. If, it were so, then $B$ would know it. In the program this fact is asserted to the situation B.A which holds the facts that $A$ knows that $B$ knows. $A$ continues reasoning: "If $B$ knew that the number on my forehead were 1, would everything be as it is now? Would it cause any contradiction?" So $A$ tries to find a contradiction in the facts that he knows that $B$ knows. From his previous answer $A$ knows that $B$ does not know what the number on his ($B$'s) forehead is. So $A$ tries to prove that $B$ would know the number on his ($B$'s) forehead, if the number on $A$'s forehead were 1, and reach to a contradiction. $B$ would know what the number on his ($B$'s)forehead is using the rule mentioned in the previous paragraph (Figure 5.7). This kind of reasoning using incoherencies in situations is performed by the constraint in Figure 5.8.

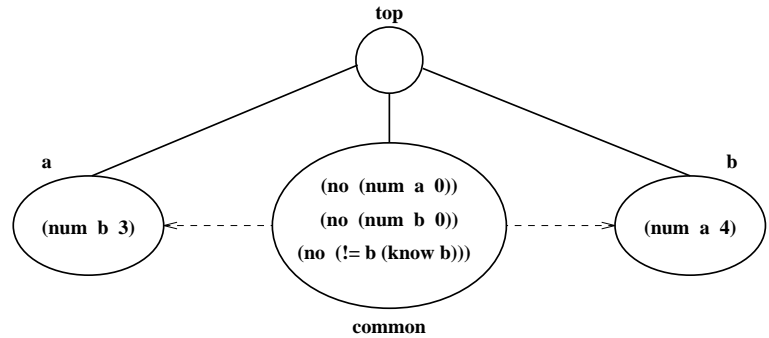

Figure 5.9: The situation tree shows the facts that $A$ knows, $B$ knows, and those that are common.

To elucidate the way the program deduces the facts about the number on the forehead of a logician, we examine in detail the situation in which `a` has 4 on his forehead, and `b` has 3. In the beginning, it is common knowledge that none of the logicians have the number 0 on their foreheads:

    (!= common (no (num a 0)))
    (!= common (no (num b 0)))

The situation tree is illustrated in Figure 5.9 where `a` and `b` are the situations that support the facts known by $A$ and $B$, and `common` denotes the situation supporting the facts that are common to both agents. The dashed arrows indicate that both `a` and `b` are inheriting the infons supported by `common`, i.e., both agents are aware of the facts that are common, and know that these facts are common.

After $B$ says "I don't know my number," the fact that the number on $A$'s forehead is not 0, turns out to be common knowledge (Figure 5.10):

    (!= common (no (num a 1)))

Next, $A$ says "I don't know my number," which means that the number on $B$'s forehead is neither 1 nor 2 (Figure 5.11):

    (!= common (no (num b 1)))
    (!= common (no (num b 2)))

Then, $B$ once again says "I don't know my number," and it is concluded that the number on $A$'s forehead is neither 2 nor 3 (Figure 5.12):

    (!= common (no (num a 2)))
    (!= common (no (num a 3)))

At this moment, $A$ deduces the fact that the number on his forehead is 4, because he knows the facts that the numbers are consecutive, that $B$'s number is 3, and that the number on his own forehead is not 2 (Figure 5.13):

    (!= a (num a 4))

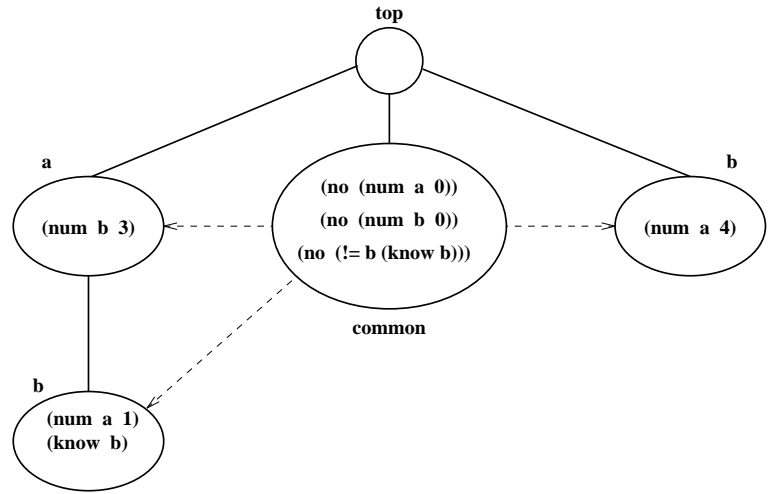

Figure 5.10: *A* makes the assumption that the number on his forehead is 1, and reaches to an incoherency.

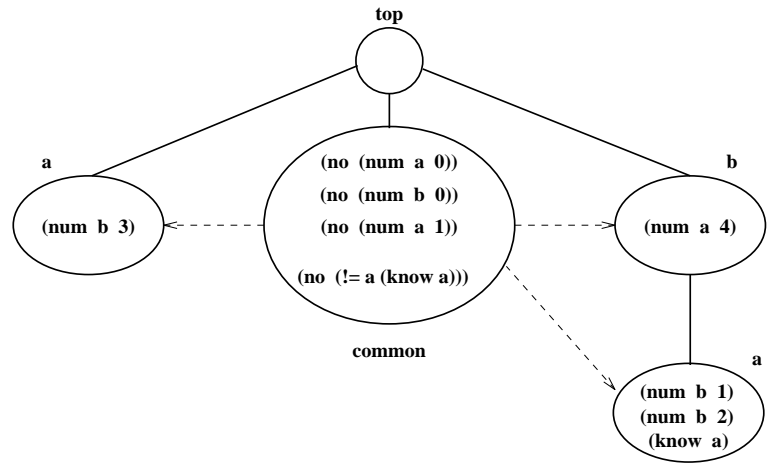

Figure 5.11: *B* makes the assumption that the number on his forehead is 1 or 2, and each time is led to an incoherency.

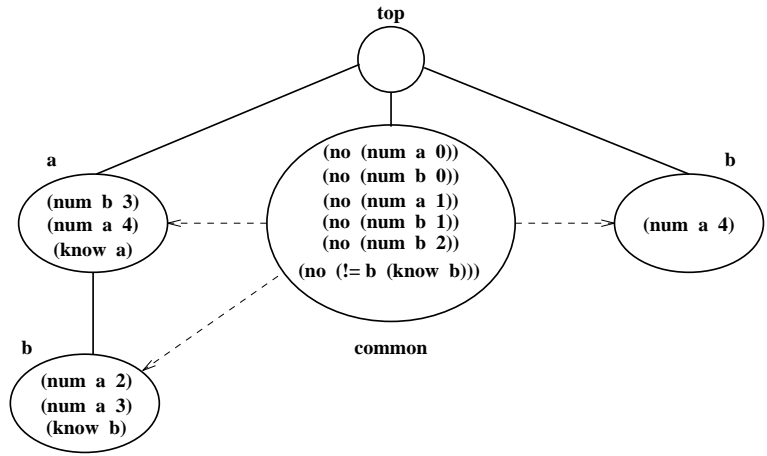

Figure 5.12: *A* finds out that the number on his forehead is neither 2 nor 3.

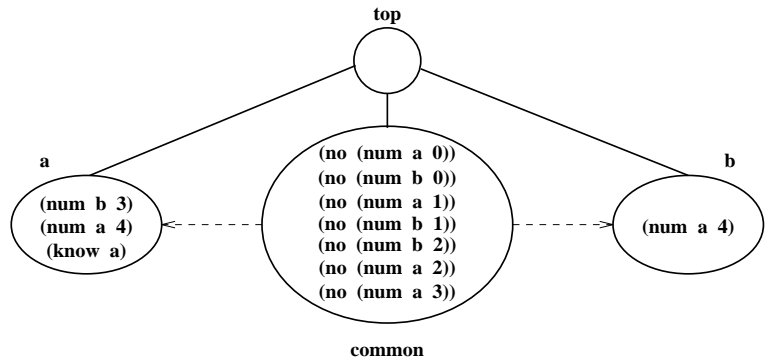

Figure 5.13: *A* knows that the number on his forehead is 4.

Note that the logicians are making intelligent assumptions. If it is known that the number on the forehead of $A$ is not $n$, then $B$ assumes that the number on his forehead is $n + 1$. At the instant when it is known that the number on the forehead of $A$ is not 0 and 1, $B$ assumes that the number on his forehead is 1 or 2, which helps him reach an incoherency, and derive new facts. However, $B$'s assuming that the number on his forehead is, say 8, would not help him much.

# Chapter 6

# Conclusion and Future Work

This paper should be considered as a field test on computational systems based on situation theory [5, 8, 14, 15, 36, 52, 53]. Our primary aim was to analyze a language (PROSIT [32, 33, 35]) which is based on situation theory and investigate applications that can grow upon the tools situation theory provides. We chose epistemic puzzles [38, 39, 40, 42, 34, 19] as the test domain, because these puzzles mainly study the knowledge individuals are aware of and the way they reason about it. We believe that situation theory provides an ontologically adequate framework to represent such puzzles.

Our results show that situation theoretic languages [32, 33, 35, 44, 48, 46, 50, 9, 10] are suitable means for human-like reasoning. PROSIT is especially appropriate for problems involving knowledge and belief. PROSIT provides some of the situation theoretic concepts such as self-referential expressions, and situations as arguments of infons. Moreover, it offers additional tools such as the situation tree hierarchy and the inheritance mechanism, which make it easier to represent individual and common knowledge.

This paper is only an initial study on 'real life' situation-theoretic applications. We hope that this study will provide the necessary motivation for further investigation on the computational systems based on situation theory and their deployment to model assorted problems of knowledge representation. A logical next step in this regard will be to exercise the capabilities of BABY-SIT [44, 48, 46].

# Appendix A

# System Predicates in PROSIT

Here, we give a brief definition of some of the system predicates in PROSIT. All of these are called by querying which can be done either at the top level (`TOP`) or within a situation. Some system predicates may also be called by asserting them, but this is only useful if the predicate has some side-effect, like printing something or asserting something into a situation.

The following describes the effects and results of the system predicates when queried in a situation *s* that is the interpreter's current focus of attention, also called the "current situation."

## A.1 Predicates for traversing the situation tree

(`!=` *sit infon*)

Asserting (`!=` *sit infon*) goes into *sit* (as described in the current situation *s*) and asserts *infon* there. Asserting (`!=` *sit infon*) causes the query (`!=` *sit infon*) to succeed thereafter (unless later retracted). (`!=` *sit infon*) expressions are not just system queries, but also infons that can be supported by situations.

(`IN` *sit*)

Focuses the interpreter on *sit*, as described in the current situation *s*. Pushes *s* onto a stack of situations. Remains in effect until an (`OUT`) or until the user-input loop is `exit`ed.

(`OUT`)

Focuses the interpreter's attention onto the topmost situation on the stack formed by previous (`IN` *sit*)'s. The interpreter thus returns to the situation that we were focused on before the `IN` that brought us here.

## A.2 Database control predicates

(`!` *infon1* ... *infonk*)

When either asserted or queried, asserts *infon1* ... *infonk* (which could be a system predicate like `!=`, `<-`, `resp`, `rule`, or `[_]`) in the current situation *s*. This may trigger the application of forward-chaining rules.

Each *infon* is always queried before being asserted (note that backward chaining rules may be used), and the assertion will not take place if the query succeeds.

(`-!` *form*)

Removes from the current situation *s* all infons *infon* matching *form* that have been explicitly asserted to be supported in *s*. Implicitly, *infon* will also be removed from the supersituations of *s*, and (!= *s infon*) will be removed from any superchunks of *s* and from the situation describing *s*.

## A.3  Constraints

(<= *head goal1 ... goaln*)

This is the form for backward-chaining constraints suitable as an argument to RESP or RULE. However, if a <= infon is asserted in a situation, it automatically becomes respected by that situation. So, in the current version there is no difference between asserting (RULE (<= ...)) and (<= ...).

(RESP *sit constr*)

When asserted, causes *sit*, as described in the current situation *s*, to respect the constraint *constr*, which may be either a backward-chaining constraint (see above) or a forward-chaining constraint (see below).

(=> *head result1 ... resultn*)

This is the form for forward-chaining constraints suitable as an argument to RESP or RULE. However, if a => infon is asserted in a situation, it automatically becomes respected by that situation. So, in the current version there is no difference between asserting (RULE (=> ...)) and (=> ...).

(RULE *constr*)

When asserted, causes the current situation *s* to respect the constraint *constr*, which may be either a backward-chaining or a forward-chaining constraint. (See above.)

When queried, succeeds if *s* has been asserted to respect *constr*.

## A.4  Control predicates, logical connectives

(AND *infon1 ... infonk*)

Identical to (! *infon1 ... infonk*) when asserted.

When queried, succeeds if and only if all of *infon1 ... infonk* succeed when queried, in order. If any one fails, or if the entire AND is backtracked to, we backtrack through the *infoni*'s until we find another solution or there are no more.

(OR *infon1 ... infonk*)

Succeeds if any of *infon1 ... infonk* succeed when queried. First tries *infon1*; if it fails, or if the entire OR is backtracked to and there are no more solutions to *infon1*, goes on to *infon2*, and so on.

(NOT *goal*)

Succeeds if and only if the given goal cannot be proven. Not the same as (NO *goal*), which asks if we *can prove* the *opposite* of *goal*.

(CUT)

Succeeds. But if it is backtracked to, it not only fails but prevents further backtracking, causing the higher-level goal (of which it is a part) to fail.

(FAIL)

Never succeeds.

(TRUE)

Always succeeds.

## A.5 Relations between situations

(@< *sit1 sit2*)

When asserted, causes the current situation *s* to describe that *sit2* supports *i* for every infon *i* valid in *sit1* and that *sit2* respects every constraint that is respected by *sit1*, i.e. *sit1* becomes a subtype of *sit2*.

(<- *sit1 sit2*)

The same as (@< *sit1 sit2*) except that only infons, but no constraints are inherited.

([_ *sit1 sit2*)

When asserted, causes *sit1* to be a subchunk of *sit2*. This means that *sit1* is totally described by the infons in *sit2*. If (!= *sit1 infon*) holds in *sit2*, then *infon* holds in *sit1*. And vice versa if *infon* holds in *sit1* then (!= *sit1 infon*) holds in *sit2*.

## A.6 Unification

(= *expr1 expr2*)

Tries to P-unify *expr1* and *expr2*. This is used both to assign a parameter to another expression (number, string, list) and to bind two unassigned parameters together so that they can be used interchangeably.

## A.7 Using Lisp within PROSIT

(LISP *expr*)

Substitutes the variables in expression *expr* with their bindings. If the result is a valid Lisp function call, then calls on Lisp to evaluate it. Succeeds if the Lisp call returns non-nil. If a variable is unbound it is used as is.

(BIND-LISP *var expression*)

If *var* is a free variable, evaluates expression as a Lisp function (as in the `lisp` predicate above) and binds *var* to the value returned by the function.

## A.8 Interacting with PROSIT

(RUN)

Starts PROSIT from Lisp. May be queried within PROSIT to create a nested, clean, empty sub-session. (Does not free stack space.)

(LOAD *filename*)

If *filename* is a string naming an existing file, reads PROSIT expressions in from the file as if the user typed them, but does not print any output. Just as in user input, the file may contain ?'s and !'s to switch between assertion and query modes, and it may contain (in *sit*) and (out) instructions so that the file gets loaded into its own situation.

(DEMO *filename*)

If *filename* is a string naming an existing file, reads PROSIT expressions from the file one at a time as if the user typed them, but waits for the user to hit return before executing each one. Useful for demonstrating sequences of queries that show off one's programs.

`(PRINTSIT)`
   Prints out each of the infons, other than `!=` or `<-` infons, which have been directly asserted in the current situation as opposed to being inherited from subsituations or subtypes.

`(TRACE)`
   Puts PROSIT into a mode where a running trace is displayed of all queries, showing when queries are called, when they are exited successfully, when they fail, and when they are backtracked to in backward-chaining.

`(DUALS)`
   Puts PROSIT into a mode wherein for each user query *query*, both the *query* and its dual (i.e., (`no` *query*)) are evaluated. If just the dual succeeds, PROSIT answers "`no`"; if both the query and its dual succeed, PROSIT answers "`yes and no`"; otherwise PROSIT answers "`yes`" or "`unknown.`"

`EXIT`
   Exits PROSIT.